\documentclass[runningheads]{llncs}
\usepackage{graphicx}
\usepackage{comment}
\usepackage{textcomp}
\usepackage{xcolor}
\usepackage{array}

\begin{document}
\title{Unsupervised Visual Time-Series Representation Learning and Clustering}

\author{Gaurangi Anand\inst{1,2} \and
Richi Nayak\inst{1,3}}

\institute{School of Computer Science, Queensland University of Technology \and
\email{gaurangianand@hdr.qut.ed.au}\and
\email{r.nayak@qut.edu.au}}

\maketitle             
\begin{abstract}
Time-series data is generated ubiquitously from Internet-of-Things (IoT) infrastructure, connected and wearable devices, remote sensing, autonomous driving research and, audio-video communications, in enormous volumes. This paper investigates the potential of unsupervised representation learning for these time-series. In this paper, we use a novel data transformation along with novel unsupervised learning regime to transfer the learning from other domains to time-series where the former have extensive models heavily trained on very large labelled datasets. We conduct extensive experiments to demonstrate the potential of the proposed approach through time-series clustering. 
\end{abstract}
\section{Introduction}\label{sec:intro}
\begin{figure}
    \centering
    \includegraphics[width=0.85\textwidth]{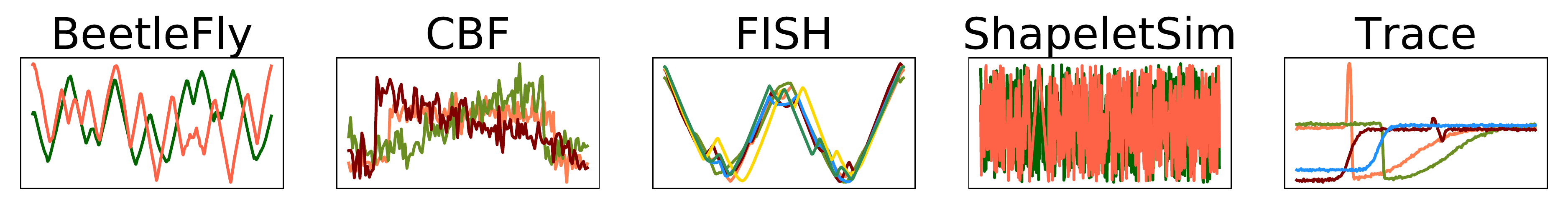}
    \caption{Five sample datasets from UCR~\cite{chen2015ucr} with one sample per cluster.}
    \label{fig:samples}
\end{figure}

Time-series data is generated ubiquitously in enormous amounts from sensors and IoT. With the prior knowledge about labels, it can be observed that the samples have minor to major shape variations across pre-defined labels, as seen in Figure~\ref{fig:samples}. These variations can be incorporated in feature representation and learning. However, a large-scale time-series labelling is expensive and requires domain expertise, paving way for unsupervised tasks for time-series~\cite{ma2019learning}. 

Time-series clustering is an unsupervised task with data representation and a suitable similarity measure as its core components. Some methods propose sophisticated similarity measures applied to raw time-series data, while others propose effective representations suitable to simpler similarity measures like Euclidean Distance (ED). Both the approaches define the means for unsupervised estimation of the extent of resemblance between the samples utilizing characteristics such as their shape~\cite{paparrizos2017fast}, shapelets~\cite{franceschi2019unsupervised}, alignment~\cite{berndt1994using} and structure~\cite{malinowski20131d}. This extent is guided by the information acquired through supervised or unsupervised learning. In the absence of labels, transfer learning (TL) is one of the solutions~\cite{bengio2012deep} where the model learned on source task, rich with labelled information, is fine-tuned for a target task. Typically the source and target datasets are related with the latter being comparatively small. The recent success of deep learning has brought advancements to TL methods in the fields of Natural Language Processing and Computer Vision (CV) with large datasets like \emph{Wikipedia}, and \emph{ImageNet}~\cite{krizhevsky2012imagenet} for fine-tuning various related data-scarce tasks. However, no such labelled time-series corpus of a similar scale exists. 

In this paper, we propose a novel approach of leveraging the large-scale training from a popular CV-based dataset~\cite{deng2009imagenet} to the time-series data in an unsupervised manner. The relatedness of ImageNet data and time-series has not been explored earlier, providing an opportunity to use visual recognition for time-series as commonly done for images. It facilitates representing time-series from a visual perspective inspired by human visual cognition, involving 2-D convolutions, unlike the 1-D convolutional approaches popular in time-series domain~\cite{franceschi2019unsupervised}. We utilize the shifted variants of the original time-series to train a model to isolate the distinct local patterns of the time-series irrespective of their location within the overall layout, and evaluate the approach for time-series clustering. More specifically, the contributions of this paper are as follows:
\begin{enumerate}
    \item We approach the problem of time-series analysis as the human visual cognitive process by transforming the 1-D time-series data into 2-D images.
    \item We leverage the training of the very large dataset available in the CV field to the unsupervised representation learning for time-series through a pre-trained 2-D deep Convolutional Neural Network (CNN) model. 
    \item We propose a novel unsupervised learning regime using triplet loss for time-series data using 2-D convolutions and shift-invariance.
    \item The resultant time-series representation has fixed length regardless of the variable length time-series that enables pairwise comparison of time-series data in linear time using Euclidean distance. 
\end{enumerate}

\section{Related Literature}\label{sec:LR}
With the availability of unlabelled datasets, several unsupervised learning methods have emerged ranging from conventional algorithms~\cite{zhang2018salient,aghabozorgi2015time} to stacked deep architectures~\cite{fawaz2019deep,ma2019learning}, for obtaining a representation based on the end task. These representations can be data adaptive like Symbolic Aggregate Approximation (SAX)~\cite{malinowski20131d} or non-data adaptive like Discrete Fourier Transform (DFT), Discrete Wavelet Transform (DWT) and Piecewise Aggregate Approximations (PAA). A model-based representation is generated by identifying the model parameters through training based on the relevant properties. Once trained, a model is used as a feature extractor~\cite{franceschi2019unsupervised} for the end task.

Similarity PreservIng RepresentAtion Learning (SPIRAL)~\cite{lei2019similarity} preserves the similarity of Dynamic Time Warping (DTW) distance in the approximated similarity matrix, to be then used for conventional partitional clustering like \emph{k}-means. \emph{k}-shape~\cite{paparrizos2017fast} adapts the clustering algorithm to the distance measure for cross-correlation while assigning clusters. Unsupervised Salient Subsequence Learning (USSL)~\cite{zhang2018salient} identifies the salient subsequences from the time-series and performs clustering based on shapelet learning and spectral analysis. Deep Embedding for Clustering (DEC)~\cite{xie2016unsupervised} and Improved Deep Embedding Clustering (IDEC)~\cite{guo2017improved} are deep learning based non-time-series clustering. Unsupervised triplet loss training has been proposed for time-series~\cite{franceschi2019unsupervised} where representations are learned and evaluated for time-series classification using 1-D dilated causal convolutions. Deep Temporal Clustering Representation (DTCR)~\cite{ma2019learning} is a deep learning-based end-to-end clustering for time-series clustering that jointly learns time-series representations and assigns cluster labels using \emph{k}-means loss.

Existing deep learning-based time-series representation methods do not use pre-trained networks due to 1) the latter not being tailored to time-series and 2) the popularity of 1-D convolutions for sequential data. Bridging this gap by utilizing a 2-D pre-trained CNN not only helps to leverage large-scale training but also provides a local pattern-based time-series representation. To the best of our knowledge, this is the first time-series representation approach combining visual perception with unsupervised triplet loss training using 2-D convolutions.

\section{Proposed Approach}\label{sec:PA}

We propose to first transform time-series into images to leverage the large-scale training from a 2-D deep CNN pre-trained with ImageNet. This CNN is then modified and re-trained for feature extraction in unsupervised setting utilizing a novel triplet loss training.   

\paragraph{\textbf{1-D to 2-D feature transformation ($f_{T}$)}}\label{sec:PA:1}
The 1-D time-series dataset is transformed into 2-D image dataset, achieved by simply \textit{plotting} a 1D time-series as a time-series image, and used as 2D matrix to take advantage of 2-D convolution operations~\cite{krizhevsky2012imagenet,he2016deep} through pre-trained 2-D CNNs. This emulation of human visual cognition to inspect/cluster time-series data using this transformation enables the vision-inspired systems to interpret the time-series visually. 
\paragraph{\textbf{CNN Architecture}}\label{sec:PA:2}
We use ResNet50~\cite{he2016deep} as our pre-trained CNN, trained on the large-scale ImageNet dataset~\cite{deng2009imagenet} for the task of object classification. It consists of residual connections that add depth to the vanilla CNN architecture~\cite{krizhevsky2012imagenet} useful for highly complex CV tasks. ImageNet~\cite{deng2009imagenet} comprises millions of images with $1000$ classes. CNN learned on this data provides visual local pattern based representations extracted as 3-D feature tensors from each of its composite layers. These feature tensors, called \textit{feature maps}, have the local spatial pattern mapping with respect to each of the input images. ImageNet allows the network to learn lots of variations in local shape patterns. With that as a premise, we argue that when using visual representations of 1-D time-series most of their shape patterns would be easily represented through a pre-trained 2-D CNN. The subsequent fine-tuning helps to add the time-series bias to shape patterns. As a result, relevant shape patterns are obtained for 2-D time-series matrices.

\begin{figure}
\centering
\includegraphics[width=0.95\textwidth]{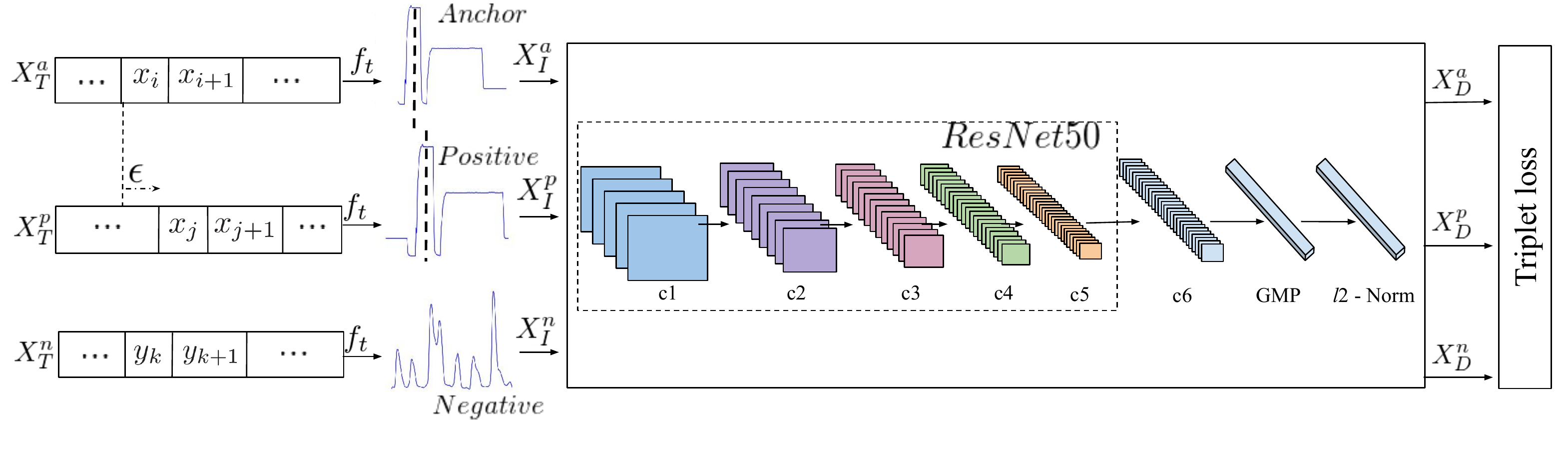}
\caption{Overview of the proposed learning framework with $c1...6$ convolution layers of the CNN followed with GMP and \textit{l}2-normalization layer.}
\label{fig:f_learn}
\end{figure}

The modified ResNet is depicted in Figure~\ref{fig:f_learn}. ResNet is retained up to its last convolution layer by removing the fully connected layer trained for the object classification task. The local spatial patterns in the form of activations for each of the time-series images are retrieved within the feature maps. A 2-D convolution layer is then appended to it, followed by the Global Max Pooling (GMP) layer~\cite{tolias2015particular}. This layer leverages the local patterns within different feature maps irrespective of their actual activation location in an image for introducing shift-invariance. It helps matching time-series images where the observed local patterns may not exactly align spatially. We then append a \textit{l}2-normalization layer to improve the network training by providing stability to it. 

Let $X_{T}$ be the set of 1-D univariate time-series that is transformed into an equivalent set of 2-D time-series images, $X_{I}$. $X_D$ refers to the representations obtained from this deep network. Using the \textit{l}2-normalized representation for each individual time-series, we generate unsupervised triplets as described below. Once trained, the \textit{l}2-normalized output is used as the Learned Time-series Representation, called as Learned Deep Visual Representation (LDVR).

\paragraph{\textbf{Triplet Selection}}\label{sec:PA:3}
Triplet loss training helps to ensure that similar instances obtain similar representations, while the dissimilar ones are pushed apart with a margin, $\alpha$. Unlike the usual supervision-based triplet selection~\cite{schroff2015facenet,franceschi2019unsupervised}, we propose a novel unsupervised triplet selection for 2-D time-series images. A positive pair consists of an original instance with its shifted variant, and its pairing with any other sample forms the negative pair. This training helps identifying the patterns that can be isolated from the layout and modify the representations such that the \textit{anchor} and \textit{positive} are brought closer and separated from \textit{negative}. 

We address the challenge of unsupervised triplet selection by considering a pool of time-series across a large number of datasets. From this pool, two time-series samples, $X_{T_a}$ and $X_{T_n}$ as anchor and negative respectively, are selected randomly from a dataset and a random shift to $X_{T_a}$ is added by introducing a circular shift of randomly chosen $\epsilon$ time-steps to obtain $X_{T_p}$, called positive. The feature transformation is applied to the triplet set to obtain $X_{I_a}$, $X_{T_p}$, and $X_{I_n}$. Figure~\ref{fig:f_learn} depicts the process of triplet selection with $X_{T_a}$, $X_{T_p}$ and $X_{T_n}$ as the triplets, where $X_{T_p}$ is obtained by introducing a shift of $\epsilon$ to $X_{T_a}$.

\paragraph{\textbf{Clustering}}\label{sec:PA:4}

The modified ResNet trained with the unsupervised Triplet loss produces a fixed length representation, $X_D$, called LDVR for each time-series image. Assuming the compact feature representation to be separable in Euclidean space, a distance based clustering algorithm likes \emph{k}-means can be applied. 

\section{Experiments}\label{sec:exp}

Extensive experiments are conducted to evaluate the accuracy of the LDVR generated by the proposed approach for the time-series clustering. 

We use the publicly available time-series UCR repository~\cite{chen2015ucr} with $85$ datasets. The sequences in each dataset have equal length ranging from $24$ to $2709$, with $2$ to $60$ classes, split into train and test. The already assigned class labels are treated as cluster identifiers and only used in evaluation.

\paragraph{\textbf{Experimental setup}}  
We use the previous version of UCR of $47$ datasets for unsupervised selection of triplets. To prevent memorization and overfitting of the network training, we randomly select $100$  samples from them\footnote{Datasets that do not contain enough samples, we chose $60\%$ of the total samples.}. For each iteration, we perform $10$ circular shifts each time with randomly chosen $\epsilon$ varying between $0.6$ and $1.0$, representing the percentage of time-series length. A total of $3056$ distinct instances were added to the pool for triplet selection. The inclusion of shifted samples forces the network to learn local patterns isolated from the layout, and use them for matching independent of their actual location across samples. All the time-series of variable length were transformed into images of $640 \times 480$ resolution. This value was derived from UCR where 80$\%$ of datasets have length below 640; to keep aspect ratio as standard 4:3, height set to 480. 

The batch-size and learning rate for training was $32$ and $0.05$ respectively, trained for $200$ epochs with filter size of $3 \times 3$ with the stride of $1$ and margin $\alpha = 1.0$ . The pre-trained \emph{ResNet50} network was frozen up to $c5$ layer and randomly initialized for next layers for triplet loss training. The best number of feature maps for $c6$ layer was estimated to be $4096$. Once trained, the representation of dimension $d = 4096$ is obtained on which clustering is performed. Figure~\ref{fig:sensitivity} (left) shows the sensitivity of LDVR with varying margin values and number of c6 feature maps w.r.t. average NMI scores.

\paragraph{\textbf{Benchmarking methods}}
We benchmarked the LDVR for time-series clustering (i.e. k-means with euclidean distance) with several methods. First, we include the Pre-trained Deep Visual Representation (PDVR) i.e., the representations extracted from the c5 layer before the Triplet loss training. We use deep learning based two non time-series clustering methods, DEC~\cite{xie2016unsupervised} and IDEC~\cite{guo2017improved} to cluster 2-D matrix time-series image data. Three traditional distance measure based methods include \emph{SPIRAL}~\cite{lei2019similarity}, k-shape ~\cite{paparrizos2017fast} and ED with k-means (ED) applied on raw time-series. Additionally, we include the two common time series representation methods, SAX (1d-SAX)~\cite{malinowski20131d} and DWT~\cite{popivanov2002similarity}. 

Two recent time-series clustering approaches: DTCR~\cite{ma2019learning} and USSL~\cite{zhang2018salient} have partial results published and have not made their source-codes available for them to be included in benchmarking. We provide a short analysis in the end.

All methods are evaluated with Normalized Mutual Information (NMI) and Rand Index (RI), the most common metrics to evaluate time-series clustering.

The network training is done on SGI Altix XE cluster with an Intel(R) Xeon(R) 2.50 GHz E5-2680V3 processor running SUSE \emph{Linux}. We use Keras $2.2.2$ with Tensorflow backend and Python 3.5. 

\section{Results: Clustering Accuracy Performance}\label{sec:result}
We first compute the cumulative ranking for both the NMI and RI scores, with the lower rank indicative of a better score. For each dataset, LDVR, PDVR and other methods are ranked based on their performance from 1 to 9, considering ties. The total rank across all datasets is computed for each method cumulatively and then averaged across all datasets, as shown in Table~\ref{tab:avg_NMI_RI_85}. 

\begin{table}[ht]
\caption{Average cumulative ranks and average scores for NMI and RI for $85$ datasets.}
\centering
\begin{tabular}{l|ccccccccc}
\hline
 & LDVR\hspace{0.12cm} & PDVR\hspace{0.12cm} & DEC\hspace{0.12cm} & IDEC\hspace{0.12cm} & SPIRAL\hspace{0.12cm} & ED\hspace{0.12cm} & k-shape\hspace{0.12cm} & SAX\hspace{0.05cm} & DWT \\ \hline
Avg. Rank (NMI)\hspace{0.1cm} & \textbf{2.74} & 3.36 & 5.25 & 5.66 & \textbf{2.66} & 3.08 & 3.98 & 3.81 & 3.12 \\ 
Avg. Score (NMI)\hspace{0.1cm} & \textbf{0.36} & 0.32 & 0.23 & 0.19 & 0.30 & 0.29 & 0.25 & 0.29 & 0.29 \\
Avg. Rank (RI)\hspace{0.1cm} & \textbf{2.42} & 3.02 & 4.66 & 5.16 & 3.09 & 3.35 & 4.22 & 3.51 & 3.40 \\ 
Avg. Score (RI)\hspace{0.1cm} & \textbf{0.73} & 0.72 & 0.66 & 0.60 & 0.69 & 0.69 & 0.63 & 0.70 & 0.69 \\ \hline
\end{tabular}
\label{tab:avg_NMI_RI_85}
\end{table}

As seen in Table~\ref{tab:avg_NMI_RI_85}, the average cumulative NMI ranks for LDVR and SPIRAL are lowest, however LDVR gets maximum wins and the maximum average NMI score, by winning in $28$ datasets. SPIRAL and PDVR win on $17$ and $16$ datasets. Moreover, the high average RI score and lowest cumulative average RI rank indicate the promising performance of LDVR, with winning on $26$ datasets for RI, followed by PDVR and SPIRAL with $18$ and $14$ wins, respectively. 

The poor results of DEC and IDEC indicate that 2-D matrix image data is not sufficient for effective clustering. PDVR is the pre-trained representation, which when fine-tuned to LDVR, helps in achieving a performance boost in more than $50\%$ of the datasets. This validates that the unsupervised fine-tuning proposed in this paper adds a bias thereby improving the clustering.

\begin{table}[ht]
\caption{Performance comparison (Average RI Score) of LDVR, DTCR and USSL}
\centering
 \begin{tabular}{cccccc}
\hline

 \vspace{0.1cm}$M_1$(\#36)\hspace{0.2cm} & $M_2$(\#36)\hspace{0.2cm} & $M_3$(\#36)\hspace{0.2cm} & \textbf{$M_3$(\#Top 36)}\hspace{0.2cm} & $M_3$(\#Top 73)\hspace{0.2cm} & $M_3$(\#All 85)\hspace{0.2cm} \\\hline
0.77 & 0.76 & 0.70 & \textbf{0.90} & 0.77 & 0.73\\\hline
\end{tabular}
\vspace{3pt}
\begin{tabular}{ccccc}
\scriptsize{\textbf{M1}: DTCR~\cite{ma2019learning}} & \hspace{1cm} & \scriptsize{\textbf{M2}: USSL~\cite{zhang2018salient}} & \hspace{1cm}
\scriptsize{\textbf{M3}: LDVR}
 \end{tabular}
 \label{sec:tableResults}
\end{table}

Table~\ref{sec:tableResults} shows the comparative performance of LDVR with DTCR~\cite{ma2019learning} and USSL~\cite{zhang2018salient}. In the absence of their source code, we compare first on their published $36$ UCR datasets in the first three columns, the reason for selection of these datasets is undisclosed. It can be observed that LDVR comes very close to DTCR and USSL. However, the average RI values on LDVR's top $36$ datasets is significantly higher than that published for the two methods. The next columns show it takes as high as $73$ UCR datasets to maintain the RI score of $0.77$, and drops only to $0.73$ when considering all $85$ datasets.  

Though many methods have been proposed for time-series analysis, shape-based clustering still appears to be challenging. The combined performance evaluation through NMI and RI establishes LDVR as the leading state-of-the-art method capable of attaining accurate time-Series representation and clusters while considering the shape of the time-series.

\subsection{Comparative Performance}
\begin{figure}
\centering
\begin{tabular}{@{}c@{}c@{}@{}c@{}c@{}c@{}}
    \includegraphics[width=0.2\textwidth]{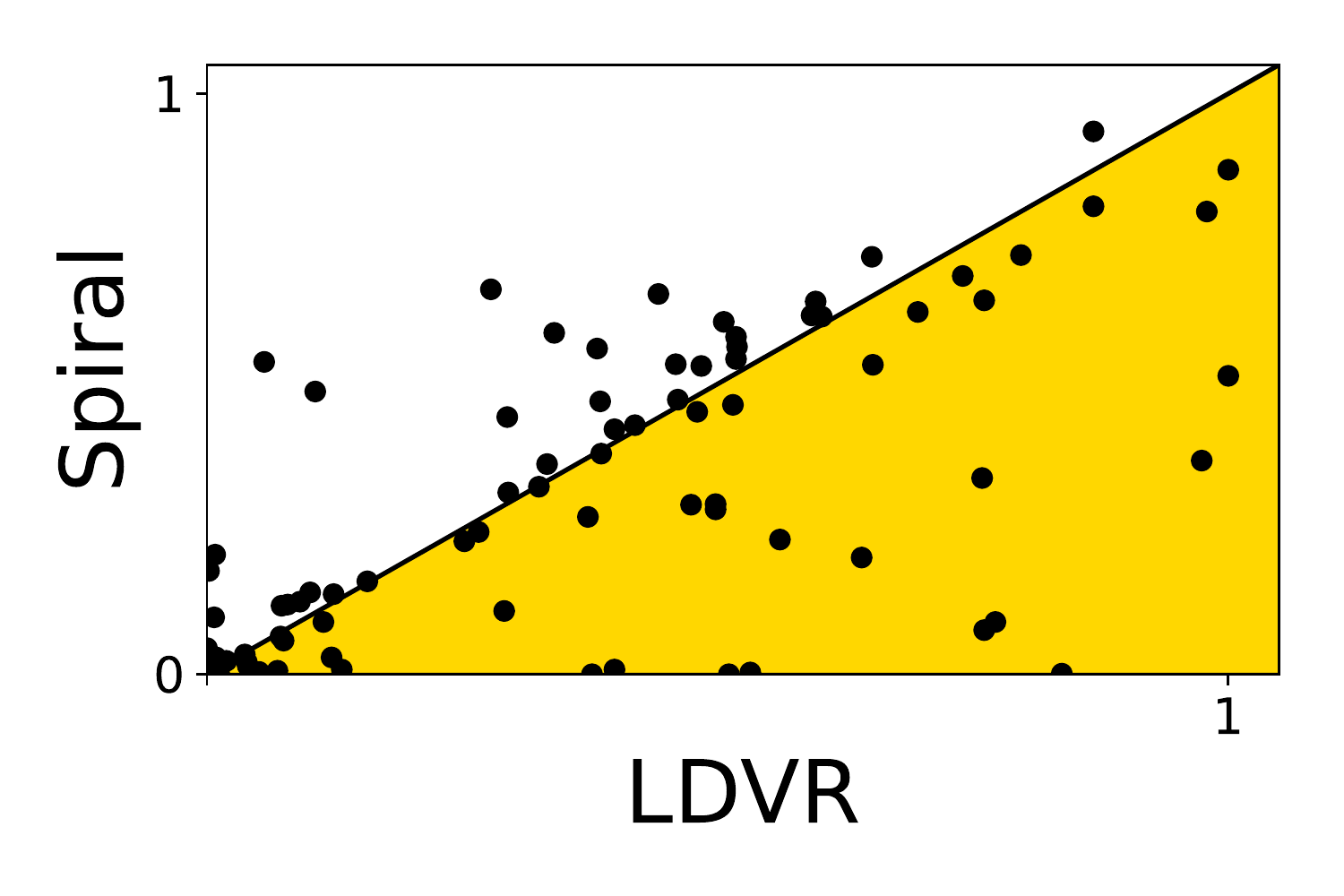} &
    \includegraphics[width=0.2\textwidth]{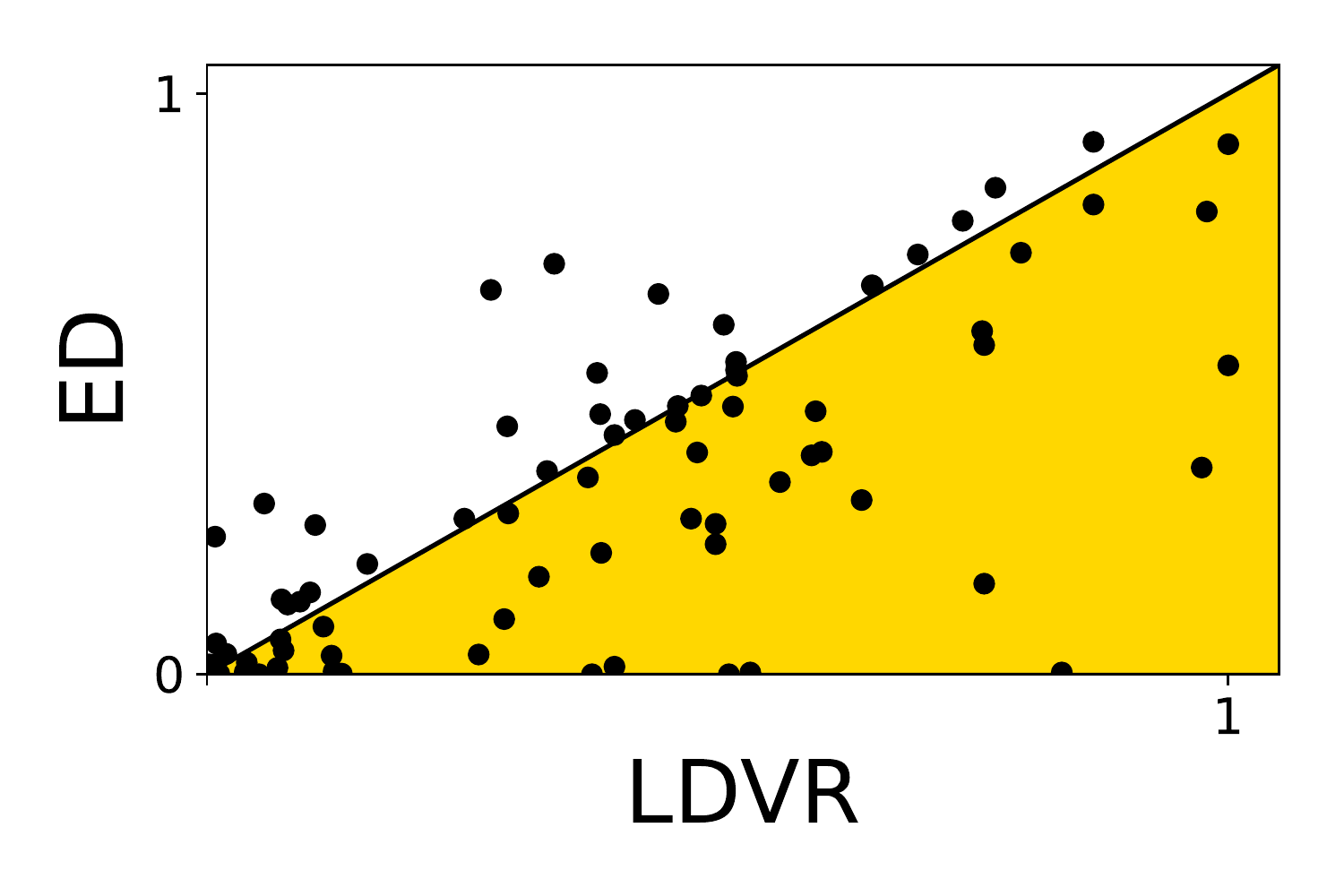}&
    \includegraphics[width=0.2\textwidth]{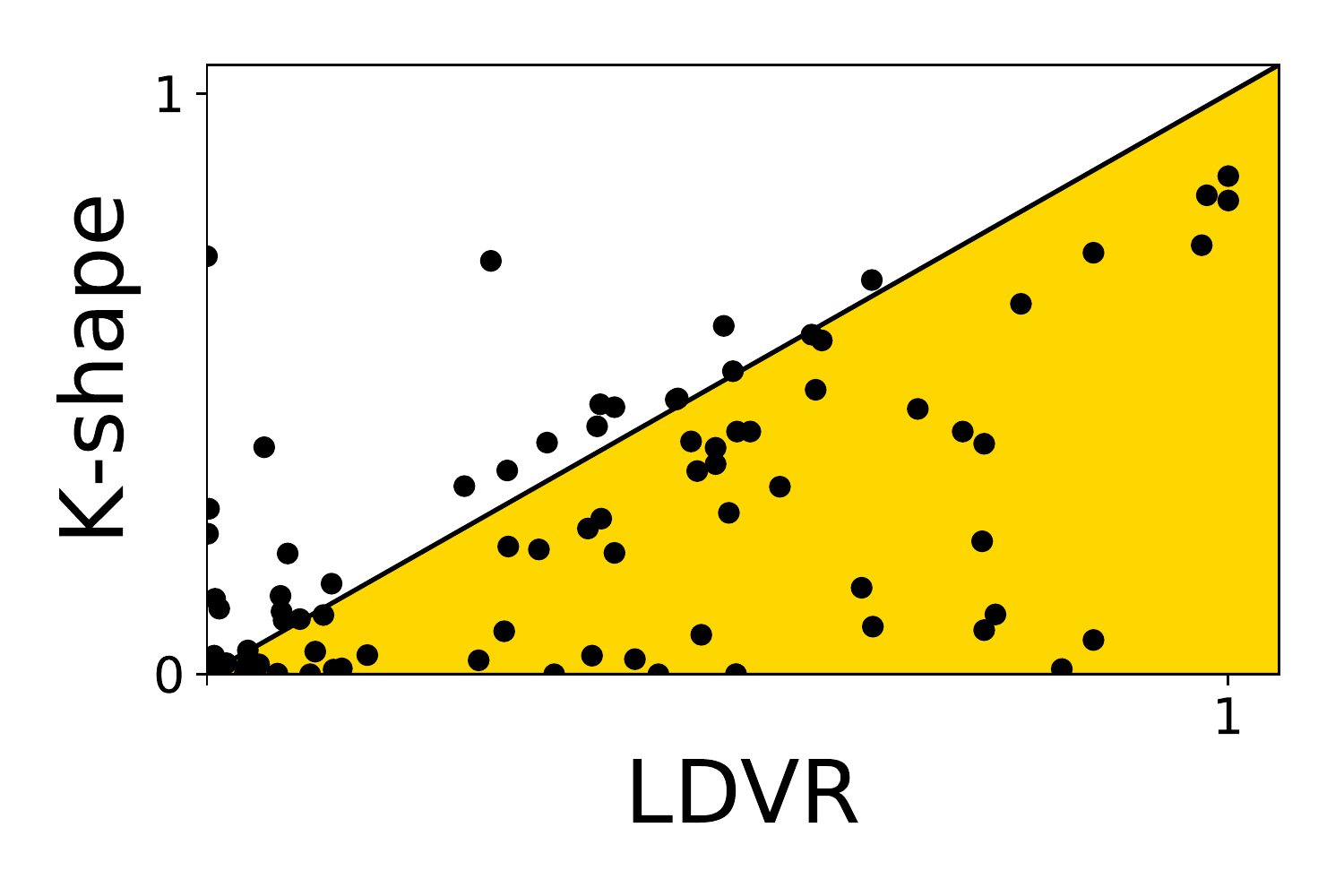} &
    \includegraphics[width=0.2\textwidth]{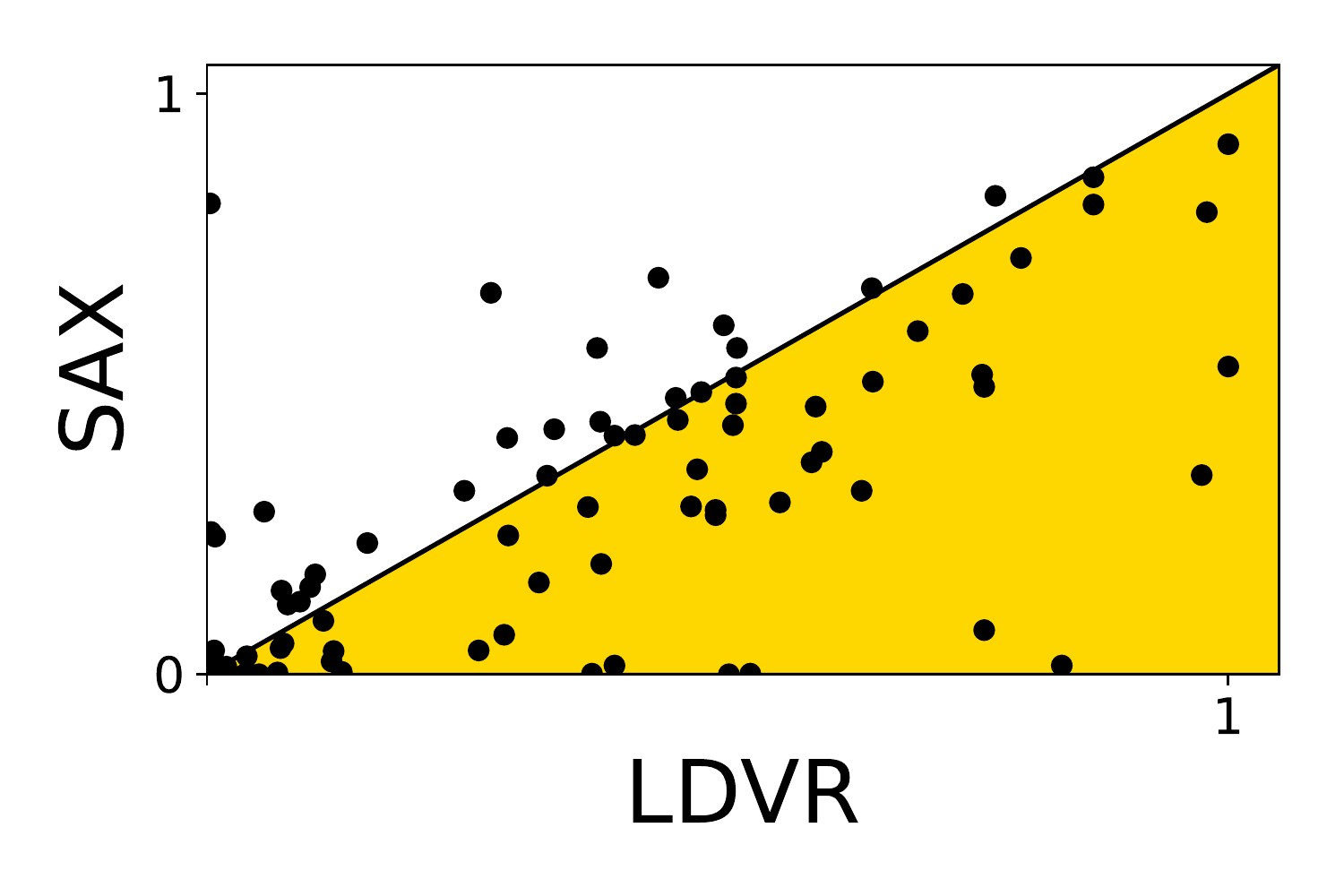} &
    \includegraphics[width=0.2\textwidth]{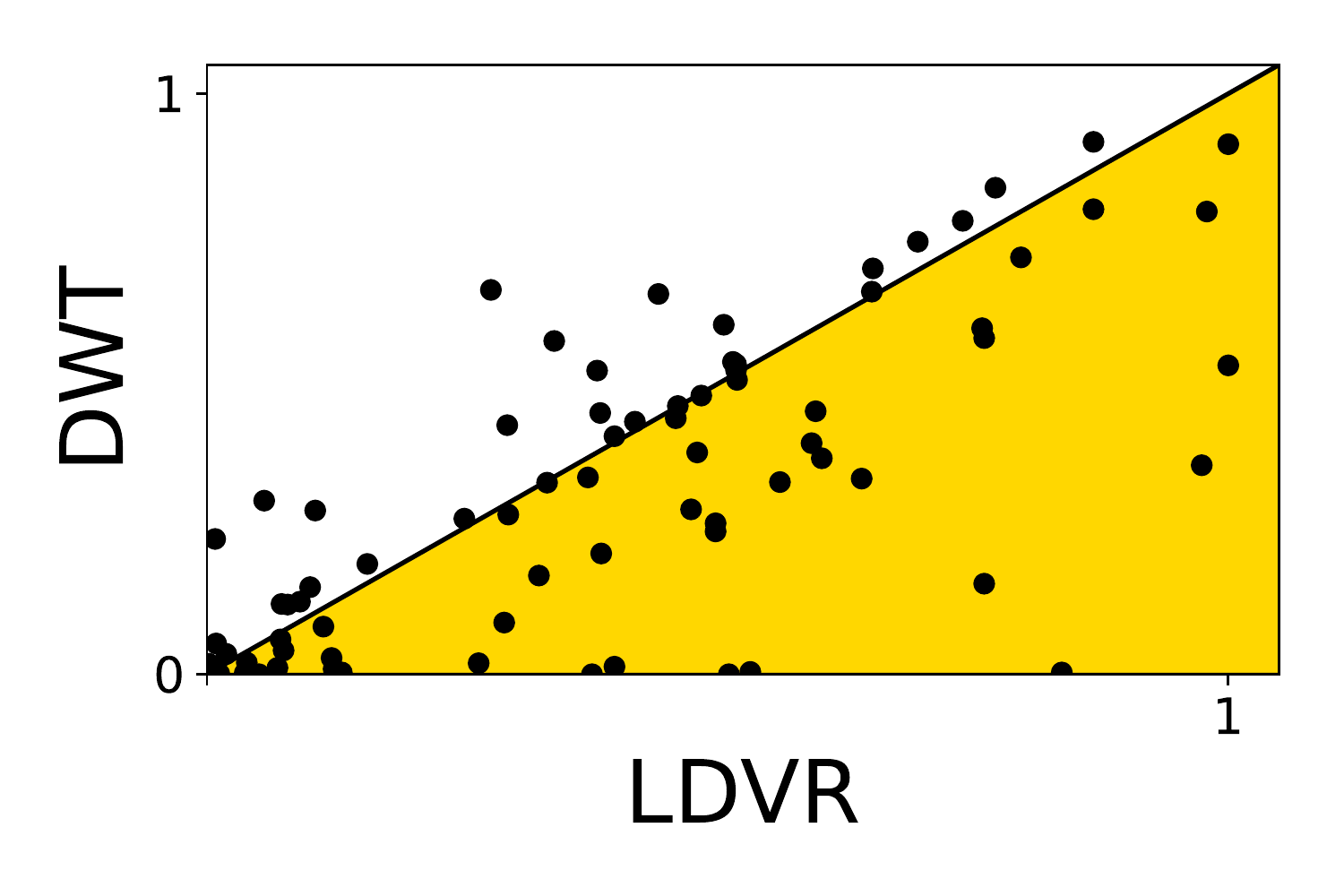}
\end{tabular}

\caption{Pairwise performance comparison of LDVR against traditional methods, with each dot as a dataset. The dots appearing in yellow triangle indicate LDVR performs better for that dataset.}

\label{fig:NMI-comparison2}
\end{figure}

Figure \ref{fig:NMI-comparison2} shows the performance of distance-measure and representation-based methods used for benchmarking compared against LDVR using NMI scores on all $85$ datasets. The larger numbers of points in yellow triangle indicate the goodness of LDVR against all of its competitors. The comparison with the leading algorithm, k-shape, ascertains that not many datasets require cross-correlation distance to be used as a shape similarity measure, with LDVR winning on $57$ out of $85$ datasets. Similarly, DTW based SPIRAL is not as effective and accurate as LDVR, losing on total $50$ datasets. 

These methods end up having false matches due to the forced alignment of time-series which is not always needed. On the other hand, k-means provide better clusters on LDVR versus ED, proving that the learned visual representations are stronger and accurate for point-to-point comparison than the raw 1-D time-series values. 

\begin{figure}
    \centering
    \includegraphics[width=0.9\textwidth]{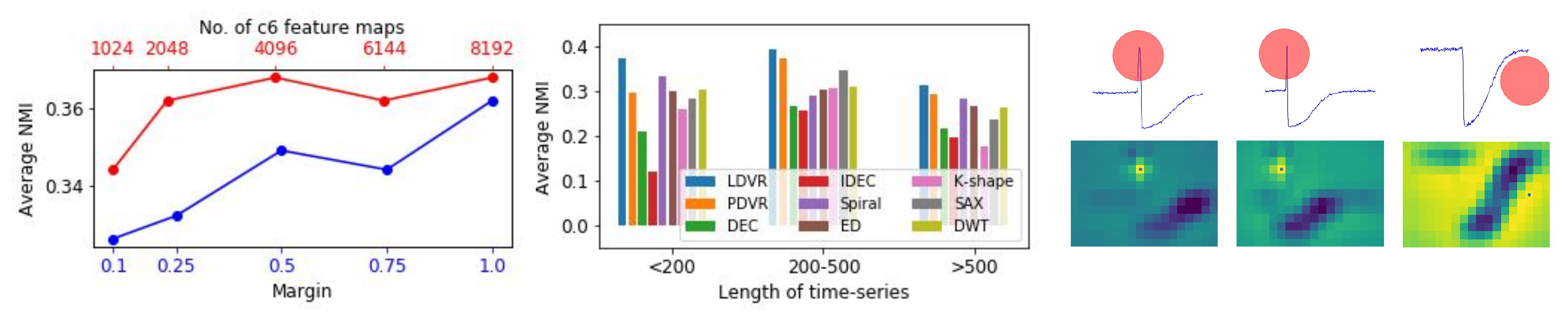}
    \caption{Left: Sensitivity analysis with respect to number of c6 feature maps (red) and margin for triplet loss (blue). Middle: Variations in average NMI performance with respect to different time-series length bins. Right: Visualization of c6 feature map activations and local feature patterns that led to correct time-series matching.}
    \label{fig:sensitivity}
\end{figure}

We also report the performance of all the methods with respect to the varying length time-series of the UCR repository. We group the UCR datasets into 3 categories based on the sequence lengths; (1) $31$ datasets with length $< 200$, (2) $29$ datasets with length $200$ to $500$, and (3) $25$ datasets with length $>500$. It can be observed from Figure~\ref{fig:sensitivity} (middle), that LVDR performs best in all the categories and does not depend on the time-series length. Despite the variable length of input time-series, the feature representation vector size in LDVR remains constant, depending on the number of feature maps in c6 layer, that is, $4096$. 

\paragraph{\textbf{Qualitative analysis with Visualization}}
Figure~\ref{fig:sensitivity} (right most) shows a triplet from the Trace dataset from UCR. The first 2 instances are anchor and positive, with the third one as negative. The discriminant pattern (marked as red in top row), observed in anchor, is slightly shifted in positive pair. The corresponding feature maps (in bottom row) show similar activations for the anchor-positive pair. However, the discriminant pattern is not seen in the negative instance which belongs to a different cluster. As a result, the feature maps corresponding to the absent patterns have different activation distribution. This proves how the local pattern based visual representation is useful and helps in identifying patterns which are similar within instances of the same cluster and discriminative across instances of different clusters.

\section{Conclusion}
We propose a novel way of generating representations for time-series using visual perception and unsupervised triplet loss training. The proposed approach has explored the relatedness of CV datasets to the time-series where the latter is expressed as 2-D image matrices, unlike the usual 1-D manner. Additionally, we utilize an existing pre-trained 2-D CNN with certain modifications to obtain an effective time-series representation, LDVR. Extensive experiments demonstrated that LDVR outperforms existing clustering methods without requiring dataset-specific training. This establishes LDVR as the leading state-of-the-art model-based representation learning method, suitable for time-series clustering. Neither being fixed to any pre-trained network, nor tied to downstream task of clustering or classification, this makes the proposed approach highly flexible and adaptable to any target task.
\bibliographystyle{splncs04}
\bibliography{mybib}

\end{document}